\documentclass{article}

\usepackage[final]{corl_2026} 

\title{What Makes a VLA Reactive? A Study of Execution Strategies for Dynamic Manipulation}
\author{
  \textbf{Zhuoyue Li\textsuperscript{1}, Zhendong Han\textsuperscript{1}, Pankhuri Vanjani\textsuperscript{1},} \\
  \textbf{Wenting Li\textsuperscript{1}, Rudolf Lioutikov\textsuperscript{1,2}} \\[4pt]
  {\small
    \textsuperscript{1}Intuitive Robots Lab (IRL), Karlsruhe Institute of Technology (KIT)
  } \\
  {\small
    \textsuperscript{2}Robotics Institute Germany (RIG)
  }
}

\usepackage{multirow}
\usepackage{graphicx}
\usepackage{booktabs}
\usepackage{amsmath}
\usepackage{cleveref}
\newcommand{\pihalf}{\ensuremath{\pi_{0.5}}}

\usepackage{hyperref}
\usepackage{amssymb}
\usepackage{subcaption}
\usepackage{makecell}
\usepackage{wrapfig}
\usepackage{tikz}
\usepackage{pgfplots}
\usepgfplotslibrary{groupplots}
\usetikzlibrary{patterns,arrows.meta}
\pgfplotsset{compat=1.18}
\definecolor{syncblue}{HTML}{1F77B4}
\definecolor{asyncorange}{HTML}{FF7F0E}
\newcommand{\cmark}{\checkmark}
\usepackage{tabularx}
\newcolumntype{L}[1]{>{\raggedright\arraybackslash}p{#1}}
\newcolumntype{Y}{>{\raggedright\arraybackslash}X}
\newcommand{\findingsfont}{\fontsize{7.5}{8.5}\selectfont}
\usepackage{placeins}  
\begin{document}
\maketitle


\begin{abstract}
Vision-language-action (VLA) models predict chunks of future actions, and when objects move, those chunks go stale before they are executed. We study what makes a pretrained VLA reactive, using \pihalf{} on two dynamic simulation benchmarks and three real-robot tasks. We vary the prediction and execution horizons, compare synchronous and asynchronous inference, test methods that deliver actions sooner or keep successive chunks continuous, and adapt the policy with motion cues and latent-space reinforcement learning. The best prediction horizon is about one second, and the best execution horizon depends on the checkpoint. Short horizons help more with asynchronous execution, and methods that keep the trajectory continuous outperform those that only reduce latency. Adapting the policy repairs targeted failures, but on the conveyor task changing the execution horizon gained as much. Reactivity depends on the execution pipeline as much as on the policy, so both should be evaluated together.
\end{abstract}


\section{Introduction}

\begin{figure}[h]
    \centering
    \includegraphics[width=\linewidth]{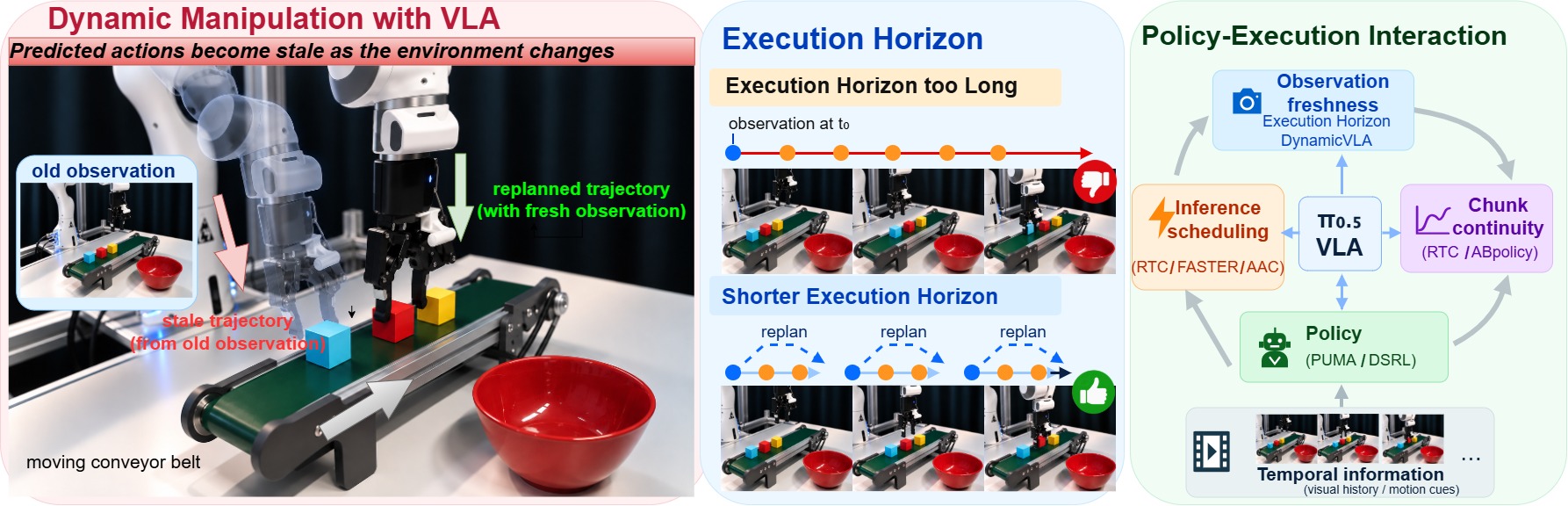}
    \caption{Dynamic VLA manipulation depends on the interaction between prediction and execution horizons, inference scheduling, chunk continuity, and policy behavior, which together determine whether frequent replanning improves task performance.}
    \label{fig:teaser}
\end{figure}
Vision-Language-Action (VLA) models  predict action chunks to support coherent motion with fewer inference calls. 
When objects move during inference and execution, these predictions can become stale. Shorter execution horizons allow more frequent replanning, but increase inference demand and can introduce discontinuities between chunks. Dynamic manipulation therefore requires balancing observation freshness, computational cost, and execution stability.
Recent approaches address these challenges through faster action generation, asynchronous execution, trajectory continuity, and temporal information~\cite{black2025rtc,lu2026fasterrethinkingrealtimeflow,xie2026dynamicvla,fang2026domino}. We used $\pi_{0.5}$ ~\cite{black2025pi05} on DOM~\cite{xie2026dynamicvla}, DOMINO~\cite{fang2026domino}, and three real-robot dynamic tasks. In this work, we investigate: \textbf{(1)} How do prediction and execution horizons affect performance? \textbf{(2)} How do scheduling and trajectory continuity affect the benefits of frequent replanning? \textbf{(3)} Can temporal information and reward-guided adaptation, including DSRL~\cite{wagenmaker2025steering}, improve dynamic manipulation? Our results show that frequent replanning alone does not ensure better performance: its benefits depend on the execution pipeline and the learned policy.

\section{Study Design}

\textbf{Problem Setup}
Given visual observations $o_t$, language instruction $\ell$, and proprioceptive state $p_t$, the policy predicts an action chunk $A_t = (a_{t\mid t}, \ldots, a_{t+H-1\mid t}) \sim \pi_\theta(\cdot \mid o_t, \ell, p_t)$, where $H$ is the \emph{prediction horizon}. The \emph{execution horizon} $s\leq H$ specifies how many actions are executed from a predicted chunk. Thus, prediction length and execution commitment can be varied separately~\cite{black2025rtc,lu2026fasterrethinkingrealtimeflow}. In \emph{synchronous execution}, the controller executes the selected actions, then waits for the next prediction. \emph{Asynchronous execution} overlaps inference with ongoing motion, reducing this waiting time. Inference latency $\tau$ corresponds to a delay of $d$ action steps. Asynchronous execution must account for motion during this delay when selecting or conditioning the next chunk. Switching between independently generated chunks can also
introduce discontinuities in the commanded trajectory~\cite{black2025rtc}. We therefore study execution horizons, inference scheduling, and continuity, and whether temporal information or policy adaptation can compensate for these effects.


\textbf{Experimental Setup}
We evaluate \pihalf{}-based configurations on DOM~\cite{xie2026dynamicvla} and DOMINO~\cite{fang2026domino} simulation tasks and three real-robot tasks using a Franka Panda: \emph{moving-cup placement}, \emph{rolling-can grasping}, and \emph{conveyor-belt picking}. The conveyor task uses three cubes with a red target cube under stationary, slow ($2.5\,\mathrm{cm/s}$), and fast
($5\,\mathrm{cm/s}$) belt conditions. Left, Middle, and Right denote the target's position along the stationary belt or its relative position within the moving three-cube sequence. 
Training, hardware, control rates, task configurations, and evaluation protocols are detailed in the appendix.

\vspace{-6pt}
\subsection{Intervention Selection and Implementation}

We select approaches that address complementary limitations of dynamic manipulation: temporal commitment, delayed or discontinuous execution, and insufficient motion information or behavioral adaptation. We implement these interventions within a common \pihalf{} framework. \Cref{tab:method_overview} summarizes their intervention points. These configurations adapt selected mechanisms from the original methods rather than reproduce every original system.

\begin{wraptable}{r}{0.6\textwidth}
\centering
\captionsetup{font=footnotesize}
\caption{Intervention levels and goal of evaluated methods.}
\label{tab:method_overview}

\scriptsize
\setlength{\tabcolsep}{3pt}
\renewcommand{\arraystretch}{1.0}
\begin{tabular}{@{}lccccl@{}}
\hline
Method & Model & Inference & Execution & Action Rep. & Target \\
\hline
RTC        &        & \cmark &        &        & Chunk transition \\
FASTER     &        & \cmark & \cmark &        & Action generation \\
DynamicVLA &        & \cmark & \cmark &        & Execution alignment \\
ABPolicy   & \cmark &        & \cmark & \cmark & Chunk continuity \\
PUMA       & \cmark &        &        &        & Motion information \\
AAC        &        & \cmark & \cmark &        & Execution horizon \\
DSRL       &        &        & \cmark &        & Latent steering \\
\hline
\end{tabular}
\end{wraptable}

\textbf{Temporal Horizons and Replanning}
The prediction horizon $H$ sets how far ahead the policy plans, and the execution horizon $s$ sets how much of that plan is carried out before the policy looks again. Together they decide how old the observation behind each executed action can be. We vary $H$ during fine-tuning and $s$ at deployment with a fixed policy. A fixed $s$ treats every moment the same, so we also test whether the policy can choose $s$ itself with Adaptive Action Chunking (AAC)~\cite{liang2026aac}. AAC samples several candidate chunks, measures the entropy of the predicted actions, and ends a chunk early when the entropy is high, subject to a minimum action-magnitude constraint in Cartesian space.

\textbf{Scheduling and Continuity Interventions}
A shorter $s$ only helps if new actions arrive in time and join the old ones smoothly. We therefore compare synchronous and asynchronous execution across execution horizons, and evaluate methods that address each requirement as separate \pihalf{} configurations: FASTER and DynamicVLA make actions available sooner, RTC and ABPolicy keep the trajectory continuous across chunks. Our DynamicVLA adaptation uses continuous inference and latency-aware action skipping, adding a minimum execution window while retaining the \pihalf{} architecture and inputs~\cite{xie2026dynamicvla}. With FASTER, we get near-term actions earlier through horizon-aware sampling, action streaming, and early stopping~\cite{lu2026fasterrethinkingrealtimeflow}. To test whether continuity-aware chunk transitions improve dynamic performance, we evaluate RTC and ABPolicy. RTC uses inference-time inpainting to align new predictions with overlapping actions from the previous chunk without additional policy training~\cite{black2025rtc}. ABPolicy combines bidirectional prediction with continuity-constrained spline refitting. We adapt it to Cartesian actions and train the policy to predict spline control points~\cite{yang2026abpolicyasynchronousbsplineflow}.

\textbf{Motion Information and Policy Adaptation}
The interventions above change how the policy is executed, the last two change the policy itself. PUMA gives it information about how the object moves, and DSRL adjusts what it does with that information. PUMA incorporates historical optical flow and future object-centric feature supervision during training~\cite{fang2026domino}. DSRL trains a latent-space actor and critic while keeping the underlying VLA frozen~\cite{wagenmaker2025steering}. We use DSRL-SAC with a short visual history aggregated by a temporal convolution for the actor and critic, while the VLA retains its current-observation input.

\section{Experimental Findings and Discussion}

We organize our evaluation around three research questions:
\textbf{RQ1:} How do prediction and execution horizons affect dynamic manipulation performance?
 \textbf{RQ2:} How do inference scheduling and trajectory continuity affect the benefits of frequent replanning?
\textbf{RQ3:} Can temporal information and reward-guided adaptation improve dynamic manipulation?

\begin{table}[!htb]
\centering
\findingsfont
\setlength{\tabcolsep}{4pt}
\renewcommand{\arraystretch}{1.1}
\caption{\textbf{RQ1: How do prediction and execution horizons affect dynamic manipulation?}}
\label{tab:rq1_findings}
\begin{tabularx}{\linewidth}{@{}L{0.30\linewidth}Y@{}}
\toprule
\textbf{Finding} & \textbf{Result and explanation} \\
\midrule
\textbf{Training:} a prediction horizon of about one second works best.
&
The best prediction horizon is 25 steps on DOM and 30 steps on DOMINO. At
their control rates of 25 and 30\,Hz, both are one second of motion
(Fig.~\ref{fig:rq1_row}a). With longer chunks, the robot keeps executing a
plan made for an object that has since moved on.
\\ \addlinespace[3pt]
\textbf{Deployment:} the best execution horizon depends on the checkpoint, not
on replanning more often.
&
The 1k checkpoint works best at $s{=}5$ and the 5k checkpoint at $s{=}15$
(Fig.~\ref{fig:rq1_row}b). The 5k policy barely moves during the first steps
of each chunk. If we replan sooner, the robot sees almost the same scene,
predicts the same slow start again, and never gets going. Replanning helps only
when the new prediction is different from the old one.
\\ \addlinespace[3pt]
\textbf{Adaptive horizons} do not help.
&
AAC never beats the best fixed horizon and fails on the static and fast belts
(Fig.~\ref{fig:rq1_row}c). It samples several candidate chunks to measure the
entropy of the predicted actions, and this extra inference delays the next
chunk. \emph{Entropy is the wrong trigger here.} It is high when the scene is
unfamiliar or when several actions are equally good, and in both cases a new
image does not change the decision. It is low when the policy is confident,
even if the object is about to move away. AAC therefore replans at the wrong
moments and pays extra inference time to do so.
\\
\bottomrule
\end{tabularx}
\end{table}
\definecolor{oiblue}{HTML}{0072B2}
\definecolor{oivermillion}{HTML}{D55E00}
\definecolor{oiorange}{HTML}{E69F00}
\definecolor{oisky}{HTML}{56B4E9}
\definecolor{oigreen}{HTML}{009E73}

\tikzset{valfont/.style={font=\fontsize{4.5}{5}\selectfont, inner sep=1pt}}
\pgfplotsset{smallbar/.style={legend image code/.code={\draw[##1, draw=none] (0cm,-0.06cm) rectangle (0.28cm,0.08cm);}}}

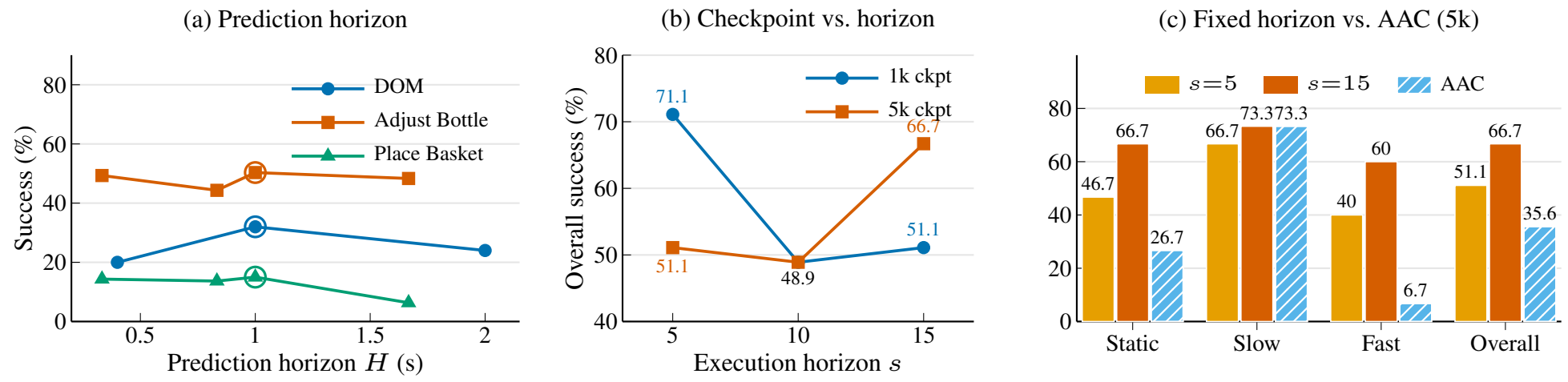
\begin{figure}[!htb]
\centering
\begin{tikzpicture}
\begin{groupplot}[
    group style={group size=3 by 1, horizontal sep=0.85cm},
    height=2.2cm,
    scale only axis,
    axis x line*=bottom,
    axis y line*=left,
    ymajorgrids=true,
    grid style={gray!20},
    tick label style={font=\fontsize{5.5}{6.5}\selectfont, inner sep=1pt},
    label style={font=\fontsize{6}{7}\selectfont},
    title style={font=\fontsize{6}{7}\selectfont, yshift=-4pt},
    every axis y label/.append style={yshift=-3pt},
    every axis x label/.append style={yshift=2pt},
    major tick length=2pt,
    tick style={black},
    clip=false,
    legend cell align=left,
    /pgf/number format/assume math mode=true
]

\nextgroupplot[
    width=3.7cm,
    title={(a) Prediction horizon},
    xlabel={Prediction horizon $H$ (s)},
    ylabel={Success (\%)},
    xmin=0.2, xmax=2.15, xtick={0.5,1,1.5,2},
    ymin=0, ymax=90, ytick={0,20,40,60,80},
    legend style={at={(0.98,0.98)}, anchor=north east, draw=none, fill=none,
        font=\fontsize{5}{6}\selectfont, row sep=-1pt},
    best/.style={only marks, mark=o, mark size=2.4pt,
        mark options={draw=#1, line width=0.6pt}, forget plot}
]
\addplot[color=oiblue, line width=0.7pt, mark=*, mark size=1.2pt]
    coordinates {(0.4,20) (1.0,32) (2.0,24)};
\addlegendentry{DOM}
\addplot[color=oivermillion, line width=0.7pt, mark=square*, mark size=1.2pt]
    coordinates {(0.333,49.33) (0.833,44.33) (1.0,50.33) (1.667,48.33)};
\addlegendentry{Adjust Bottle}
\addplot[color=oigreen, line width=0.7pt, mark=triangle*, mark size=1.4pt]
    coordinates {(0.333,14.33) (0.833,13.67) (1.0,15.00) (1.667,6.33)};
\addlegendentry{Place Basket}
\addplot[best=oiblue]       coordinates {(1.0,32)};
\addplot[best=oivermillion] coordinates {(1.0,50.33)};
\addplot[best=oigreen]      coordinates {(1.0,15.00)};

\nextgroupplot[
    width=2.9cm,
    title={(b) Checkpoint vs.\ horizon},
    xlabel={Execution horizon $s$},
    ylabel={Overall success (\%)},
    xmin=3, xmax=17, xtick={5,10,15},
    ymin=40, ymax=80, ytick={40,50,60,70,80},
    legend style={at={(1.0,1.02)}, anchor=north east, draw=none, fill=none,
        font=\fontsize{5}{6}\selectfont, row sep=-1pt}
]
\addplot[color=oiblue, line width=0.7pt, mark=*, mark size=1.2pt]
    coordinates {(5,71.1) (10,48.9) (15,51.1)};
\addlegendentry{1k ckpt}
\addplot[color=oivermillion, line width=0.7pt, mark=square*, mark size=1.2pt]
    coordinates {(5,51.1) (10,48.9) (15,66.7)};
\addlegendentry{5k ckpt}
\node[valfont, text=oiblue,       anchor=south, yshift=1.5pt]  at (axis cs:5,71.1)  {71.1};
\node[valfont, text=oiblue,       anchor=south, yshift=1.5pt]  at (axis cs:15,51.1) {51.1};
\node[valfont, text=oivermillion, anchor=north, yshift=-1.5pt] at (axis cs:5,51.1)  {51.1};
\node[valfont, text=oivermillion, anchor=south, yshift=1.5pt]  at (axis cs:15,66.7) {66.7};
\node[valfont, text=black,        anchor=north, yshift=-1pt]   at (axis cs:10,48.9) {48.9};

\nextgroupplot[
    width=4.0cm,
    title={(c) Fixed horizon vs.\ AAC (5k)},
    ybar=0.6pt,
    bar width=7.5pt,
    enlarge x limits=0.15,
    xtick={0,1,2,3},
    xticklabels={Static,Slow,Fast,Overall},
    ymin=0, ymax=100, ytick={0,20,40,60,80},
    point meta=y,
    nodes near coords={\pgfmathprintnumber[fixed, precision=1]{\pgfplotspointmeta}},
    every node near coord/.append style={valfont, text=black, anchor=south, yshift=0.5pt},
    smallbar,
    legend columns=3,
    legend style={at={(0.5,0.99)}, anchor=north, draw=none, fill=none,
        font=\fontsize{5}{6}\selectfont,
        /tikz/every even column/.append style={column sep=4pt}}
]
\addplot[fill=oiorange, draw=white, line width=0.3pt]
    coordinates {(0,46.7) (1,66.7) (2,40.0) (3,51.1)};
\addlegendentry{$s{=}5$}
\addplot[fill=oivermillion, draw=white, line width=0.3pt]
    coordinates {(0,66.7) (1,73.3) (2,60.0) (3,66.7)};
\addlegendentry{$s{=}15$}
\addplot[fill=oisky, draw=white, line width=0.3pt,
         postaction={pattern=north east lines, pattern color=white}]
    coordinates {(0,26.7) (1,73.3) (2,6.7) (3,35.6)};
\addlegendentry{AAC}

\end{groupplot}
\end{tikzpicture}
\caption{\textbf{RQ1.} (a) Success against the prediction horizon in seconds on
DOM and DOMINO; rings mark the best horizon. (b) Conveyor-belt success of the
1k and 5k checkpoints across execution horizons. (c) Success of the 5k
checkpoint with fixed horizons and AAC per belt condition ($H{=}20$,
asynchronous, 45 trials each).}
\label{fig:rq1_row}
\end{figure}

\begin{table}[!htb]
\centering
\findingsfont
\setlength{\tabcolsep}{4pt}
\renewcommand{\arraystretch}{1.1}
\caption{\textbf{RQ2: When does more frequent replanning actually help?}}
\label{tab:rq2_findings}
\begin{tabularx}{\linewidth}{@{}L{0.30\linewidth}Y@{}}
\toprule
\textbf{Finding} & \textbf{Result and explanation} \\
\midrule
\textbf{Scheduling:} short horizons help more with asynchronous execution.
&
At $s{=}10$, the synchronous \pihalf{} and PUMA policies succeed in only
$20\%$ of trials, while the asynchronous methods reach $60$--$75\%$
(Fig.~\ref{fig:rq2_row}c). On DOM the gap grows as the horizon shrinks
(Fig.~\ref{fig:rq2_row}a). In synchronous execution, the robot finishes a
chunk and then waits for the policy to compute the next one; during that wait
it does not move. A shorter horizon means more chunks, so more waits, and the
object moves on during each of them. In asynchronous execution the next chunk
is computed while the current one is still running, so a short horizon adds no
waiting.
\\ \addlinespace[3pt]
\textbf{Continuity:} methods that preserve continuity lead, and low latency
alone does not.
&
ABPolicy is the best method; RTC matches FASTER despite FASTER's lower latency.
(Fig.~\ref{fig:rq2_row}c, 
Table~\ref{tab:ttfa_hardware}).
A fresh action is only useful if the robot can follow it. When a new chunk
starts far from where the old one ended, the robot has to jump, and that costs
more than the faster inference saves.
\\ \addlinespace[3pt]
\textbf{Action rate:} on the fast belt, every method improves at 30\,Hz.
&
Raising the action rate from 20 to 30\,Hz lifts ABPolicy from $13$ to $93\%$
and FASTER from $0$ to $33\%$ on the fast belt (Fig.~\ref{fig:rq2_row}b). The
same chunk is finished sooner, so the object has less time to move away before
the next observation.
\\
\bottomrule
\end{tabularx}
\end{table}

\definecolor{syncblue}{HTML}{1F77B4}
\definecolor{asyncorange}{HTML}{FF7F0E}
\definecolor{mPUMA}{HTML}{0072B2}
\definecolor{mABP}{HTML}{E69F00}
\definecolor{mFASTER}{HTML}{CC79A7}
\definecolor{matblue}{HTML}{0072B2}

\pgfplotsset{
    syncbar/.style={
        ybar, bar width=5pt, bar shift=-2.8pt,
        fill=syncblue, draw=syncblue, mark=none,
        every node near coord/.append style={xshift=-2.8pt}
    },
    asyncbar/.style={
        ybar, bar width=5pt, bar shift=2.8pt,
        fill=asyncorange, draw=asyncorange, mark=none,
        every node near coord/.append style={xshift=2.8pt}
    },
    beltpanel/.style={
        ybar=0.6pt, bar width=6pt, enlarge x limits=0.35,
        ymin=0, ymax=110, ytick={0,20,40,60,80,100},
        point meta=y,
        nodes near coords={\pgfmathprintnumber[fixed, precision=0]{\pgfplotspointmeta}},
        every node near coord/.append style={
            font=\fontsize{4.5}{5}\selectfont, text=black,
            anchor=south, yshift=0.5pt, inner sep=0pt},
        area legend
    }
}

\begin{figure}[!htb]
\centering
\begin{tikzpicture}[baseline=(current bounding box.north)]
\begin{groupplot}[
    group style={group size=2 by 1, horizontal sep=0.8cm},
    height=2.2cm,
    scale only axis,
    axis x line*=bottom,
    axis y line*=left,
    ymajorgrids=true,
    grid style={gray!20},
    tick label style={font=\fontsize{5.5}{6.5}\selectfont, inner sep=1pt},
    label style={font=\fontsize{6}{7}\selectfont},
    title style={font=\fontsize{6}{7}\selectfont, yshift=-4pt},
    every axis y label/.append style={yshift=-3pt},
    every axis x label/.append style={yshift=2pt},
    major tick length=2pt,
    tick style={black},
    clip=false,
    legend cell align=left,
    /pgf/number format/assume math mode=true
]

\nextgroupplot[
    width=2.8cm,
    title={(a) DOM dynamic task},
    xlabel={Execution horizon $s$},
    ylabel={Success (\%)},
    xmin=-0.6, xmax=4.6,
    xtick={0,1,2,3,4}, xticklabels={2,5,10,15,25},
    ymin=0, ymax=45, ytick={0,10,20,30,40},
    point meta=y,
    nodes near coords={\pgfmathprintnumber{\pgfplotspointmeta}},
    every node near coord/.append style={
        font=\fontsize{4.5}{5}\selectfont, text=black,
        anchor=south, yshift=0.5pt, inner sep=0pt},
    area legend,
    legend style={at={(0.98,0.98)}, anchor=north east, draw=none, fill=none,
        font=\fontsize{5}{6}\selectfont, row sep=-1pt}
]
\addplot+[syncbar]  coordinates {(0,19) (1,22) (2,18) (3,19) (4,10)};
\addlegendentry{Synchronous}
\addplot+[asyncbar] coordinates {(0,32) (1,28) (2,25) (3,23) (4,12)};
\addlegendentry{Asynchronous}

\nextgroupplot[
    beltpanel,
    width=4.2cm,
    bar width=4.5pt,
    enlarge x limits=0.16,
    ymax=130, ytick={0,20,40,60,80,100},
    title={(b) Conveyor: belt condition and action rate},
    xtick={0,1,2,3}, xticklabels={Static, Slow, Fast\,20\,Hz, Fast\,30\,Hz},
    legend columns=3,
    legend style={at={(0.5,0.99)}, anchor=north, draw=none, fill=none,
        font=\fontsize{5}{6}\selectfont,
        /tikz/every even column/.append style={column sep=3pt}}
]
\addplot[fill=mPUMA,   draw=none] coordinates {(0,33.3) (1,60.0)  (2,33.3) (3,60.0)};
\addlegendentry{PUMA}
\addplot[fill=mABP,    draw=none] coordinates {(0,33.3) (1,100.0) (2,13.3) (3,93.3)};
\addlegendentry{ABPolicy}
\addplot[fill=mFASTER, draw=none] coordinates {(0,66.7) (1,100.0) (2,0)    (3,33.3)};
\addlegendentry{FASTER}
\draw[gray!60, dashed, line width=0.4pt] (axis cs:1.5,0) -- (axis cs:1.5,108);

\end{groupplot}
\end{tikzpicture}%
\hspace{0.45cm}%
\begin{tikzpicture}[baseline=(current bounding box.north),
    x=0.50cm, y=-0.36cm,
    cell/.style={minimum width=0.48cm, minimum height=0.34cm, inner sep=0pt,
        font=\fontsize{5}{5.5}\selectfont},
    rowlab/.style={anchor=east, font=\fontsize{5.5}{6.5}\selectfont, inner sep=2pt},
    collab/.style={anchor=south, font=\fontsize{5.5}{6.5}\selectfont, inner sep=1pt}
]
\node[collab, font=\fontsize{6}{7}\selectfont] at (3,-1.55) {(c) Conveyor: success (\%) vs.\ horizon $s$};
\foreach \hs [count=\ci from 0] in {5,10,15,20,25,40,50} {
    \node[collab] at (\ci,-0.5) {\hs};
}
\foreach \name/\cells [count=\ri from 0] in {
    \pihalf{} sync/{1/20,2/35,3/50,4/50,5/40,6/40},
    PUMA sync/{1/20,3/55,5/50,6/60},
    PUMA async/{1/60,2/60,4/65,5/70},
    FASTER/{0/70,1/75,2/75,4/70,5/50},
    RTC/{1/75,2/75,4/65,5/50},
    DynamicVLA/{2/55,4/60,5/55}%
} {
    \node[rowlab] at (-0.5,\ri) {\name};
    \foreach \ci in {0,...,6} { \node[cell, fill=gray!8] at (\ci,\ri) {}; }
    \foreach \ci/\val in \cells {
        \pgfmathtruncatemacro{\shade}{\val}
        \ifnum\shade>55
            \node[cell, fill=matblue!\shade!white, text=white] at (\ci,\ri) {\val};
        \else
            \node[cell, fill=matblue!\shade!white, text=black] at (\ci,\ri) {\val};
        \fi
    }
}
\node[rowlab] at (-0.5,6) {ABPolicy};
\foreach \ci in {0,...,6} { \node[cell, fill=gray!8] at (\ci,6) {}; }
\node[cell, fill=matblue!95!white, text=white] at (1,6) {95$^{*}$};
\draw[black, line width=0.5pt] (-0.3,1.5) -- (6.3,1.5);
\end{tikzpicture}
\caption{\textbf{RQ2.} (a) Success on the dynamic DOM task with the same policy
($H{=}25$) under synchronous and asynchronous execution. (b) Conveyor-belt
success of the best configuration of each method on the static, slow, and fast
belt, and on the fast belt at a higher action rate; each bar is 15 trials.
(c) Conveyor-belt success of every method across execution horizons, with
synchronous methods above the rule. Blank cells were not evaluated;
ABPolicy ($^*$) was evaluated only at $s=9$,
as its effective execution horizon depends on inference delay.}
\label{fig:rq2_row}
\end{figure}
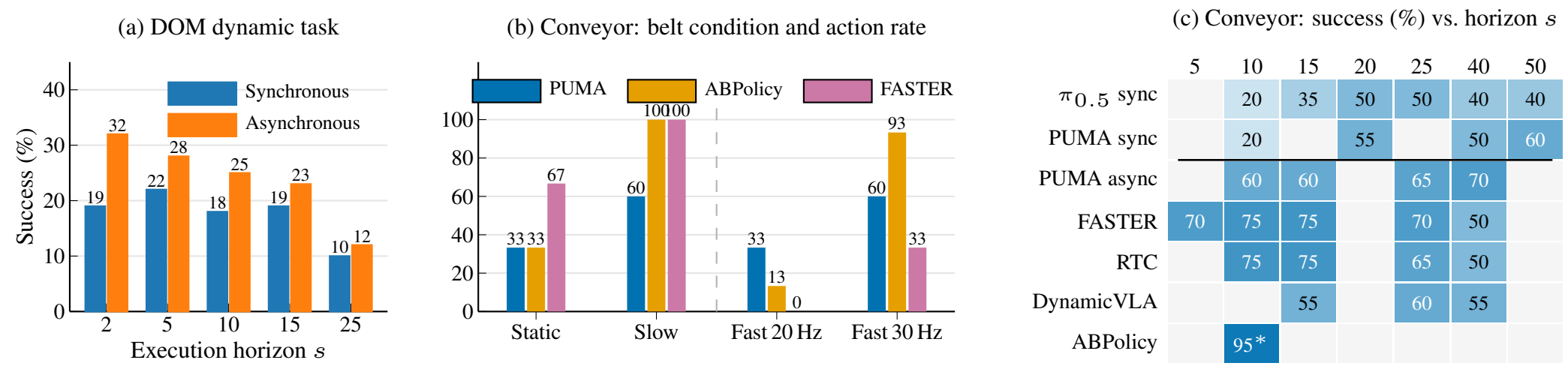

\begin{table}[!htb]
\centering
\findingsfont
\setlength{\tabcolsep}{4pt}
\renewcommand{\arraystretch}{1.1}
\caption{\textbf{RQ3: Can temporal information or reward-guided adaptation make the policy itself more reactive?}}
\label{tab:rq3_findings}
\begin{tabularx}{\linewidth}{@{}L{0.30\linewidth}Y@{}}
\toprule
\textbf{Finding} & \textbf{Result and explanation} \\
\midrule
\textbf{Motion information:} PUMA gives mixed results.
&
PUMA does better than \pihalf{} on the conveyor but worse on the rolling can
(Fig.~\ref{fig:rq3_row}a). Motion cues help on the conveyor, and asynchronous
execution makes them count: run synchronously, PUMA still fails at short
horizons like the baseline, but run asynchronously it reaches $60$--$70\%$ at
every horizon (Fig.~\ref{fig:rq2_row}c). Knowing where the object is going is
only useful when the policy also acts on fresh chunks.
\\ \addlinespace[3pt]
\textbf{Latent steering:} DSRL repairs targeted failures without touching the VLA.
&
On the targeted condition, success rises from $32$ to $47\%$ on DOM and from
$0$ to $60\%$ on the hardest conveyor case (Fig.~\ref{fig:rq3_row}b). The gain
depends on the checkpoint and disappears at a large noise magnitude. DSRL only
changes the noise that the frozen policy turns into actions, so it can reach a
behaviour only if the policy already contains it.
\\ \addlinespace[3pt]
\textbf{Scope of steering:} the gains stay local and match what a change of execution
horizon gives for free.
&
Averaged over all nine conditions, DSRL changes success by $4$ to $20$ points
and some conditions get worse (Fig.~\ref{fig:rq3_row}c). It lifts the 1k
checkpoint to $64$--$69\%$, but the same checkpoint reaches $71\%$ at $s{=}5$
with no training at all (Fig.~\ref{fig:rq1_row}b). Changing how the policy is
executed was worth as much as training it.
\\
\bottomrule
\end{tabularx}
\end{table}
\definecolor{cbase}{HTML}{56B4E9}  
\definecolor{cint}{HTML}{D55E00}   
\definecolor{cgroup}{HTML}{009E73} 

\pgfplotsset{
    rq3bars/.style={
        ybar=0.6pt, bar width=5pt,
        ymin=0, ymax=125, ytick={0,20,40,60,80,100},
        nodes near coords,
        every node near coord/.append style={
            font=\fontsize{4.5}{5}\selectfont, text=black,
            anchor=south, yshift=0.5pt, inner sep=0pt},
        area legend,
        legend style={draw=none, fill=none, font=\fontsize{5}{6}\selectfont, row sep=-1pt}
    }
}

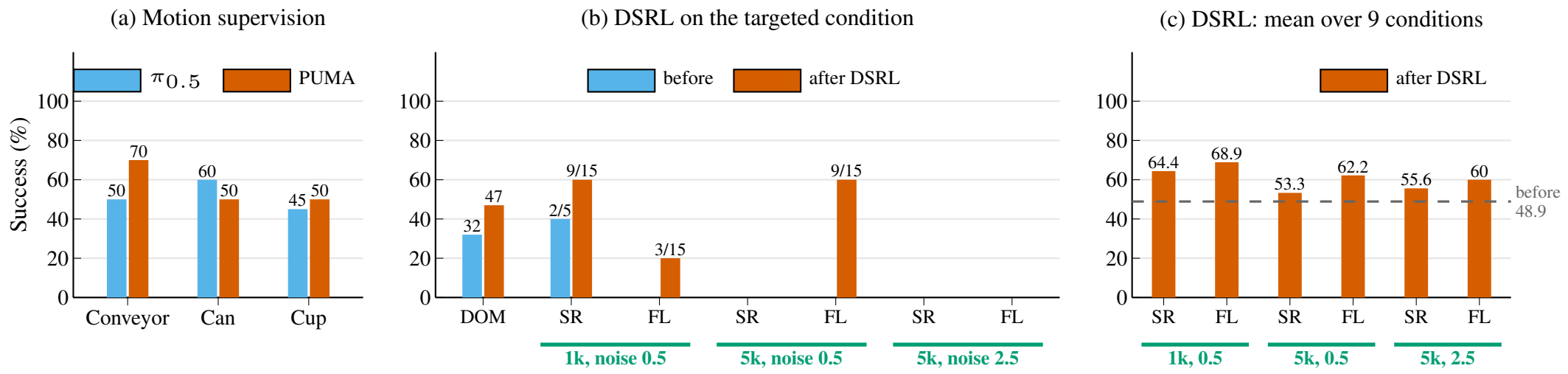
\begin{figure}[!htb]
\centering
\begin{tikzpicture}
\begin{groupplot}[
    group style={group size=3 by 1, horizontal sep=0.65cm},
    height=2.2cm,
    scale only axis,
    axis x line*=bottom,
    axis y line*=left,
    ymajorgrids=true,
    grid style={gray!20},
    tick label style={font=\fontsize{5.5}{6.5}\selectfont, inner sep=1pt},
    label style={font=\fontsize{6}{7}\selectfont},
    title style={font=\fontsize{6}{7}\selectfont, yshift=-4pt},
    every axis y label/.append style={yshift=-3pt},
    major tick length=2pt,
    tick style={black},
    clip=false,
    legend cell align=left,
    /pgf/number format/assume math mode=true
]

\nextgroupplot[
    rq3bars,
    width=2.6cm,
    enlarge x limits=0.3,
    title={(a) Motion supervision},
    ylabel={Success (\%)},
    xtick={0,1,2}, xticklabels={Conveyor, Can, Cup},
    point meta=y,
    nodes near coords={\pgfmathprintnumber[fixed, precision=0]{\pgfplotspointmeta}},
    legend columns=2,
    legend style={at={(0.5,0.99)}, anchor=north,
        /tikz/every even column/.append style={column sep=3pt}}
]
\addplot[fill=cbase, draw=none] coordinates {(0,50) (1,60) (2,45)};
\addlegendentry{\pihalf{}}
\addplot[fill=cint,  draw=none] coordinates {(0,70) (1,50) (2,50)};
\addlegendentry{PUMA}

\nextgroupplot[
    rq3bars,
    width=5.6cm,
    enlarge x limits=0.09,
    title={(b) DSRL on the targeted condition},
    xtick={0,1,2,3,4,5,6},
    xticklabels={DOM, SR, FL, SR, FL, SR, FL},
    x tick label style={font=\fontsize{5}{5.5}\selectfont, inner sep=1pt},
    point meta=explicit symbolic,
    legend columns=2,
    legend style={at={(0.5,0.99)}, anchor=north,
        /tikz/every even column/.append style={column sep=3pt}}
]
\addplot[fill=cbase, draw=none] coordinates {
    (0,32)   [32]
    (1,40)   [2/5]
    (2,0)    [~]
    (3,0)    [~]
    (4,0)    [~]
    (5,0)    [~]
    (6,0)    [~]
};
\addlegendentry{before}
\addplot[fill=cint, draw=none] coordinates {
    (0,47)   [47]
    (1,60)   [9/15]
    (2,20)   [3/15]
    (3,0)    [~]
    (4,60)   [9/15]
    (5,0)    [~]
    (6,0)    [~]
};
\addlegendentry{after DSRL}
\draw[cgroup, line width=1.0pt] (axis cs:0.65,-24) -- (axis cs:2.35,-24);
\node[font=\fontsize{5}{5.5}\selectfont\bfseries, text=cgroup, anchor=north, inner sep=1pt] at (axis cs:1.5,-25) {1k, noise 0.5};
\draw[cgroup, line width=1.0pt] (axis cs:2.65,-24) -- (axis cs:4.35,-24);
\node[font=\fontsize{5}{5.5}\selectfont\bfseries, text=cgroup, anchor=north, inner sep=1pt] at (axis cs:3.5,-25) {5k, noise 0.5};
\draw[cgroup, line width=1.0pt] (axis cs:4.65,-24) -- (axis cs:6.35,-24);
\node[font=\fontsize{5}{5.5}\selectfont\bfseries, text=cgroup, anchor=north, inner sep=1pt] at (axis cs:5.5,-25) {5k, noise 2.5};

\nextgroupplot[
    rq3bars,
    width=3.4cm,
    bar width=6pt,
    xmin=-0.5, xmax=5.5,
    title={(c) DSRL: mean over 9 conditions},
    xtick={0,1,2,3,4,5},
    xticklabels={SR, FL, SR, FL, SR, FL},
    x tick label style={font=\fontsize{5}{5.5}\selectfont, inner sep=1pt},
    point meta=y,
    nodes near coords={\pgfmathprintnumber[fixed, precision=1]{\pgfplotspointmeta}},
    legend style={at={(0.99,0.99)}, anchor=north east}
]
\addplot[fill=cint, draw=none] coordinates {(0,64.4) (1,68.9) (2,53.3) (3,62.2) (4,55.6) (5,60.0)};
\addlegendentry{after DSRL}
\draw[black!60, dashed, line width=0.6pt] (axis cs:-0.5,48.9) -- (axis cs:5.5,48.9);
\node[font=\fontsize{4.5}{5}\selectfont, text=black!60, anchor=west, inner sep=1pt, align=left]
    at (axis cs:5.5,48.9) {before\\48.9};
\draw[cgroup, line width=1.0pt] (axis cs:-0.35,-24) -- (axis cs:1.35,-24);
\node[font=\fontsize{5}{5.5}\selectfont\bfseries, text=cgroup, anchor=north, inner sep=1pt] at (axis cs:0.5,-25) {1k, 0.5};
\draw[cgroup, line width=1.0pt] (axis cs:1.65,-24) -- (axis cs:3.35,-24);
\node[font=\fontsize{5}{5.5}\selectfont\bfseries, text=cgroup, anchor=north, inner sep=1pt] at (axis cs:2.5,-25) {5k, 0.5};
\draw[cgroup, line width=1.0pt] (axis cs:3.65,-24) -- (axis cs:5.35,-24);
\node[font=\fontsize{5}{5.5}\selectfont\bfseries, text=cgroup, anchor=north, inner sep=1pt] at (axis cs:4.5,-25) {5k, 2.5};

\end{groupplot}
\end{tikzpicture}
\caption{\textbf{RQ3.} (a) Best success of \pihalf{} and PUMA
on each real-robot task. (b) Success on the targeted condition before and after
DSRL, on DOM in percent and on the conveyor belt as success counts over 5 trials
before and 15 after; brackets give the checkpoint and the noise magnitude.
(c) Mean success over all nine conveyor conditions after DSRL; the dashed line
is the $48.9\%$ of the original checkpoints.}
\label{fig:rq3_row}
\end{figure}

\FloatBarrier
\vspace{-4pt}
\section{Conclusion and Limitations}
We asked what makes a pretrained VLA reactive. In our experiments, much of the answer lies in how the policy is executed. The prediction horizon is best chosen in seconds, and the best execution horizon depends on the checkpoint. A shorter horizon only helped when the new prediction actually changed the action. Short horizons worked well when inference ran alongside execution and when successive chunks joined smoothly. Lower latency alone did not help. Motion cues and latent steering helped in targeted cases. On the conveyor task, however, a different execution horizon gained as much without any training. We suggest evaluating the execution strategy together with the policy, and reporting observation age, action availability, and chunk continuity alongside success.

Our findings characterise \pihalf{}-based adaptations of the evaluated methods on a small set of tasks. Real-robot comparisons use 15--20 trials per setting, and the can and cup tasks rely on manually induced object motion, so small differences should not be over-interpreted. Adaptive horizon strategies should weigh the expected benefit of replanning, inference overhead, and trajectory continuity, not action uncertainty alone.


\clearpage


\bibliography{reference}  
\appendix
\section*{Appendix}
\addcontentsline{toc}{section}{Appendix}

\section{Related Work}
\label{app:related_work}
\subsection{Action Chunking and Execution Horizons}
\label{app:Actionchunking}

Action chunking has become a common design choice in modern robot policies, including many Vision-Language-Action (VLA) models. Rather than predicting a single action at each timestep, a chunk-based policy predicts a sequence of future actions jointly, an idea popularized in robotic imitation learning by Action Chunking with Transformers (ACT) \cite{zhao2023learning}. This formulation provides a temporally structured action representation and reduces the need to make each control decision independently. Modern VLA systems retain this basic interface while combining it with large-scale visual, language, and robot representations. For example, $\pi_{0.5}$ uses a flow-based action expert to generate continuous robot action sequences from multimodal observations \cite{black2025pi05}.

The prediction horizon determines how many future actions are generated from a single observation, whereas the execution horizon specifies how many actions are executed before replanning. In practice, the execution horizon can be substantially shorter than the prediction horizon, allowing the controller to update its predictions using new observations \cite{black2025rtc,zhao2026dehp,lu2026fasterrethinkingrealtimeflow}. Shorter execution horizons increase replanning frequency and can improve closed-loop responsiveness, but also increase inference demand and the frequency of transitions between independently generated action chunks. The appropriate horizon can additionally depend on the action representation. 

\subsection{Real-Time Action Generation and Inference Scheduling}
\label{app:realtime}

\begin{figure*}[h]
    \centering
    \includegraphics[width=\textwidth]{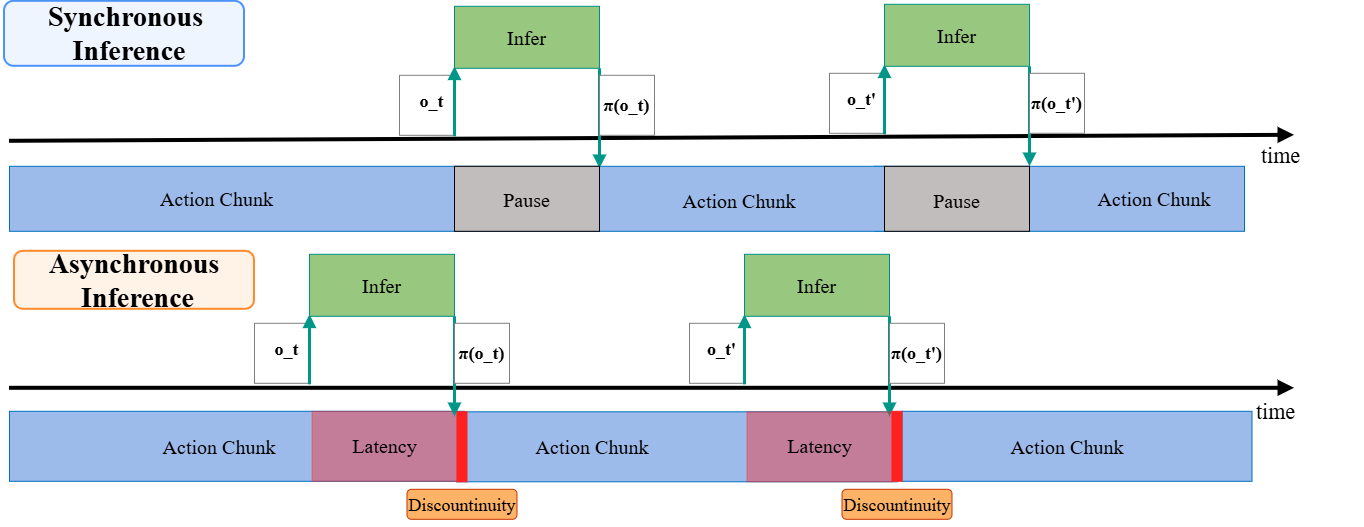}
    \caption{Synchronous execution introduces waiting periods, while asynchronous execution overlaps inference with motion but may cause inter-chunk discontinuities.}
    \label{fig:sync_async_inference}
\end{figure*}

In synchronous execution, the robot completes one action chunk before requesting the next, introducing pauses between chunks. Asynchronous execution overlaps action generation with ongoing motion, reducing idle time but introducing new timing challenges. As shown in \Cref{fig:sync_async_inference}, actions may become outdated during inference, while switching between independently generated chunks can introduce trajectory discontinuities.

DynamicVLA addresses inference delay and outdated actions through Continuous Inference (CI) and Latency-Aware Action Streaming (LAAS) \cite{xie2026dynamicvla}. CI overlaps inference and execution, while LAAS discards outdated actions and prioritizes newer predictions when chunks overlap.

FASTER shifts the focus from full-chunk completion time to \emph{time to first action} (TTFA) \cite{lu2026fasterrethinkingrealtimeflow}. Its Horizon-Aware Schedule (HAS) generates near-term actions with fewer denoising steps, allowing them to be streamed before the entire chunk is finalized. RAVEL instead uses rolling action generation and asynchronous VLM encoding to maintain partially denoised future actions and incorporate recent visual information through a lightweight observation pathway \cite{chen2026ravelasynchronousrollinginference}.

A complementary approach is IMLE-VLA, which replaces iterative action sampling with a single-step conditional generator to reduce inference latency \cite{Hosseinkhani2026IMLEVLAFS}. Beyond inference, Realtime-VLA V2 addresses sensing delays, trajectory feasibility, and task-dependent execution speed to improve physical robot execution \cite{yang2026realtimevla2}.

These methods improve action availability and temporal alignment through different mechanisms. However, faster action generation alone does not guarantee better closed-loop responsiveness, as execution timing and trajectory continuity remain important.

\subsection{Trajectory Continuity and Action Representation}
\label{app:continuity}

Asynchronous execution can introduce discontinuities when switching between independently generated action chunks. Real-Time Chunking (RTC) addresses this problem through inference-time action inpainting \cite{black2025rtc}. It uses soft masking to preserve near-term continuity while allowing future actions to adapt to new observations. RTC requires no retraining but introduces additional computation through gradient-based guidance.

Training-Time Action Conditioning extends RTC by learning prefix conditioning during training instead of applying inference-time guidance, reducing inference overhead at the cost of additional training \cite{black2025trainingtimeactionconditioningefficient}. PAINT addresses the same prefix-consistency problem through initial noise selection rather than modifying the denoising process \cite{ho2026start}. AR-VLA takes a different approach by using an autoregressive action expert with persistent action history to maintain temporal consistency across predictions \cite{hu2026arvlatrueautoregressiveaction}. These methods address temporal consistency at different levels: inference-time guidance, training-time conditioning, noise initialization, and persistent action history.

From the action-representation perspective, ABPolicy combines asynchronous inference with a B-spline control-point representation \cite{yang2026abpolicyasynchronousbsplineflow}. Cubic B-splines provide $C^2$ continuity within each chunk. To reduce inter-chunk discontinuities, ABPolicy combines Bidirectional Action Prediction (BiAP), which incorporates past actions into prediction, with Continuity-Constrained Refitting (CCR), which aligns the new trajectory with recently executed actions.

\subsection{Dynamic Manipulation and Temporal Reasoning}
\label{app:dynamic}

Dynamic manipulation introduces additional challenges because objects may move independently of the robot. A predicted action sequence can become outdated before it is fully executed, while a single visual observation provides limited information about object motion. DynamicVLA introduces the Dynamic Object Manipulation (DOM) benchmark to evaluate manipulation under independently moving objects and changing environmental conditions \cite{xie2026dynamicvla}. The DOMINO benchmark further evaluates generalization across dynamic tasks, motion patterns, and robot embodiments \cite{fang2026domino}.

Temporal alignment alone is insufficient for understanding object motion. In dynamic environments, the policy must reason not only about where an object is currently located, but also about its motion and likely future state. PUMA addresses this problem by incorporating historical optical flow and future object-centric feature prediction into the policy representation \cite{fang2026domino}. Historical optical flow provides explicit motion cues, while learnable world queries are trained to predict future object features. This encourages the policy to anticipate object motion without requiring the future-prediction branch at inference time.

\subsection{Adaptive Replanning and Policy Adaptation}
\label{app:adaptive}

A fixed execution horizon assumes that the same number of predicted actions should be executed throughout a task. In practice, different situations may require different execution horizons. Adaptive Action Chunking (AAC) uses uncertainty in the predicted action sequence to determine how long the current chunk should be executed \cite{liang2026aac}. High action uncertainty generally leads to shorter execution and earlier replanning, while low uncertainty permits longer execution.

Other approaches learn the execution horizon from task reward rather than relying on action uncertainty. Adaptive Action Chunking via Multi-Chunk Q Value Estimation (ACH) evaluates different prefixes of the same predicted chunk using learned Q-values \cite{shin2026ach}. Dynamic Execution Horizon Prediction (DEHP) instead learns a categorical policy that selects the execution horizon while keeping the underlying chunk generator frozen \cite{zhao2026dehp}. Both methods use reinforcement learning to adapt action commitment, but differ in whether they explicitly estimate the values of candidate prefixes or directly learn a horizon-selection policy.

A complementary direction is to adapt the generated behavior rather than the execution horizon. Diffusion Steering via Reinforcement Learning (DSRL) keeps a pretrained diffusion or flow generator frozen and learns to steer its sampling process through latent noise \cite{wagenmaker2025steering}. This allows reinforcement learning to select higher-reward behaviors without updating the underlying generative model. Real-Time EXPO-FT extends reinforcement-learning-based adaptation to online action correction by using a lightweight reactive policy to modify remaining actions based on updated observations \cite{dong2026reinforcementlearningrealtimevisionlanguageaction}.

\section{Implementation Details}
\label{app:implementation}
\subsection{Prediction and Execution Horizons}
We train separate policies with different prediction horizons $H$ to study the duration of the predicted trajectory. For execution-horizon comparisons, we keep the trained checkpoint and prediction horizon fixed and vary the number of actions consumed from each prediction. Checkpoint comparisons are reported separately because changing the trained policy can also change its response to frequent replanning.

\subsection{Adaptive Action Chunking}

AAC estimates action uncertainty from $N=20$ action chunks sampled in parallel and selects an execution horizon using an entropy criterion and a minimum action-magnitude constraint~\cite{liang2026aac}. Our implementation computes uncertainty over the Cartesian translation, rotation, and gripper components. Gaussian differential entropy is used for continuous actions, while categorical entropy is used for discrete gripper commands. The average entropy $\overline{E}_s$ is calculated over the first $s$ actions.

The execution horizon is selected as
\begin{equation*}
    s^* = \max\left(\arg\max_s \left(\overline{E}_{s+1}-\overline{E}_s\right), \xi\right),
\end{equation*}
where $\xi$ imposes a minimum action-magnitude constraint,
\begin{equation*}
    \xi = \min\{l \mid m(l)>\alpha\}.
\end{equation*}
Here, $m(l)$ combines accumulated translational displacement, rotational change, and gripper switching, while $\alpha$ denotes the movement threshold. In our real-robot experiments, $\alpha=0.4$ achieved the best performance among the tested values.

After selecting $s^*$, the controller executes the corresponding portion of the predicted chunk before replanning. Sampling multiple candidate chunks increases inference latency, which is included in the evaluated execution pipeline.

\subsection{DynamicVLA Execution Mechanisms}
We transfer continuous inference and latency-aware action skipping from DynamicVLA~\cite{xie2026dynamicvla} to the \pihalf{} client-server pipeline. The client updates an observation mailbox while inference is running. Each returned chunk is indexed relative to the observation used to generate it, and actions corresponding to elapsed action intervals are discarded before execution.

Our implementation retains the \pihalf{} architecture and current-observation input. It therefore differs from the complete DynamicVLA model, which uses a smaller backbone and temporal observations. We also introduce a minimum execution window $W$: a new prediction replaces the current chunk only after at least $W$ actions from the current chunk have been executed. This modification limits the frequency of chunk switching and reduces how quickly new predictions can affect the robot.

\subsection{FASTER}
We implement FASTER~\cite{lu2026fasterrethinkingrealtimeflow} using horizon-aware sampling, action streaming, prefix conditioning, and early stopping. The server sends actions as they finish denoising. The client stores the received actions by index and executes them when required. Each request includes the actions expected to be executed during inference, which form the conditioned prefix of the next prediction.

The horizon-aware schedule uses $u_0=0.9$ and $\alpha=0.6$, selected in preliminary real-robot experiments. 
Denoising stops once the conditioned prefix and the actions required for the selected execution horizon have been finalized.

We first implemented the inference-time components, then added mixed-schedule fine-tuning and training-time action conditioning. During fine-tuning, the horizon-aware and constant schedules are sampled with equal probability, $p=0.5$. Training-time conditioning provides a clean action prefix while the remaining actions are noisy, exposing the policy to the prefix structure used during execution.

\subsection{Real-Time Chunking}
The initial RTC chunk is generated synchronously to populate the action buffer. Subsequent requests include the latest observation, the remaining previous prediction, and an estimate of the inference delay~\cite{black2025rtc}.

During denoising, RTC guides the new chunk toward the overlapping actions from the previous prediction. Actions expected to execute during inference receive the strongest guidance. The guidance weight decreases over the remaining overlap, while actions beyond the overlap are generated without this constraint. The correction requires a vector-Jacobian product at each denoising step.

The execution horizon is set to
\begin{equation*}
    s=\max(d,s_{\min}),
\end{equation*}
where $s_{\min}$ is the prescribed minimum horizon and $d$ is updated from measured inference delays.

\subsection{Cartesian ABPolicy Adaptation}
ABPolicy~\cite{yang2026abpolicyasynchronousbsplineflow} predicts B-spline control points and uses continuity-constrained refitting before executing a new trajectory. We retain the Cartesian action representation of our demonstrations.

Positions are fitted directly. Orientations are expressed relative to a reference quaternion $q_{\mathrm{ref}}$,
\begin{equation*}
    q_{\mathrm{rel},t}=q_{\mathrm{ref}}^{-1}\otimes q_t,
\end{equation*}
and mapped to rotation vectors before fitting. During training, the reference is taken from the present-time boundary of the trajectory window; during inference, it is the current robot orientation. Reconstructed rotation vectors are converted back to quaternions before execution.

We use $P=8$ past actions and $H=32$ future actions, giving a 40-action trajectory represented by $N=8$ control points. During refitting, the first $N_{\mathrm{free}}=4$ control points are optimized against the recently executed actions; the remaining control points retain their predicted values. The reference contains both the $P$ past actions and the actions executed during inference. A regularization term with weight $0.05$ penalizes displacement of the last free control point from its predicted value.

\subsection{PUMA and Offline Feature Caching}
For PUMA~\cite{fang2026domino}, the client records a history of static-camera images. Four historical frames are sampled at a stride of four action steps, corresponding to $t-16$, $t-12$, $t-8$, and $t-4$. At $20\,\mathrm{Hz}$, this history spans $0.8\,\mathrm{s}$. Each historical frame is compared with the current frame to compute an optical-flow map, which is encoded alongside the current observations.

Four world queries receive supervision from object-centric features at $t+4$, $t+8$, $t+12$, and $t+16$. GroundingDINO localizes the target object, SAM2 segments it, and a frozen DINO encoder extracts the target feature. These features are generated once and stored in an offline cache. During training, the corresponding cached features supervise the world queries with a loss weight of $0.05$.

The teacher models are used only to prepare training targets. Historical optical flow is computed during inference. Our PUMA implementation uses synchronous execution by default. We additionally evaluate an asynchronous execution variant to study the effect of inference scheduling.

\subsection{DSRL with Temporal Visual History}
We use DSRL-SAC~\cite{wagenmaker2025steering} to train a latent-space actor and critic while keeping the fine-tuned VLA frozen. At each policy query, the actor selects a latent input $w_t$, which the VLA maps to an action chunk:
\begin{equation*}
    w_t\sim\pi_\phi^W(\cdot\mid z_t),
    \qquad
    A_t=G_\theta(o_t,\ell,p_t,w_t).
\end{equation*}
Here, $z_t$ contains the steering policy's proprioceptive and visual features, and $G_\theta$ is the frozen VLA generator.

For each camera, three frames separated by three action steps are encoded independently using a shared convolutional encoder. A temporal convolution aggregates their features. The latest-frame feature is projected separately and added as a residual to retain the current spatial information. At $20\,\mathrm{Hz}$, consecutive sampled frames are separated by $0.15\,\mathrm{s}$, so the history spans $0.30\,\mathrm{s}$.

The actor and critic receive this temporal representation. The underlying VLA continues to receive its current observation and the selected latent input. Real-robot training uses a binary task-success reward supplied after each rollout. The execution horizon is fixed at $s=10$ for the reported steering experiments.

\section{Experimental Setup and Evaluation Protocol}
\label{app:experimental_setup}

\subsection{Base Policy and Training}

We use \pihalf{} as the base policy and perform full supervised fine-tuning on the demonstrations for each task. Training runs on NVIDIA A100 GPUs with an effective batch size of 128 for all configurations. We use three baseline checkpoints: the 1k and 5k checkpoints are trained with a prediction horizon of $H=20$ and correspond to 1,000 and 5,000 fine-tuning steps, respectively, while the 1.5k checkpoint is trained with $H=50$ for 1,500 steps.

\subsection{Simulation Tasks}

\begin{figure*}[!t]
\centering

\includegraphics[width=0.30\textwidth]{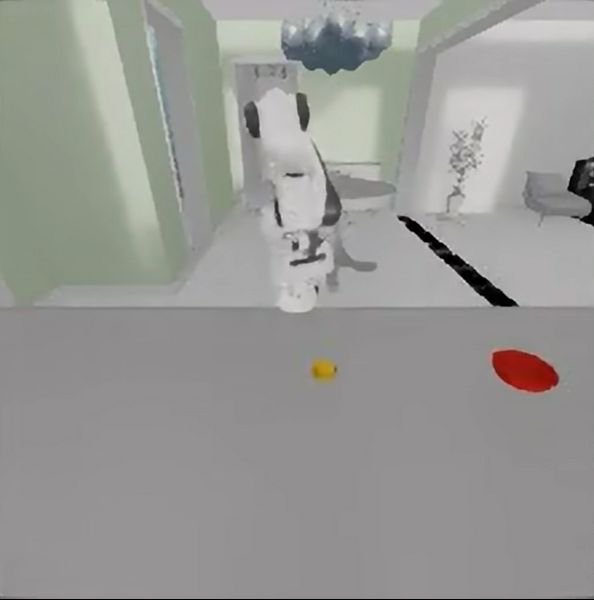}
\hfill
\includegraphics[width=0.30\textwidth]{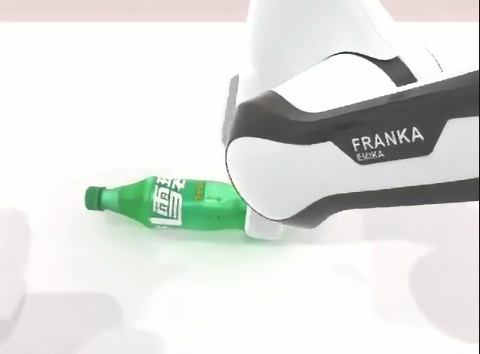}
\hfill
\includegraphics[width=0.30\textwidth]{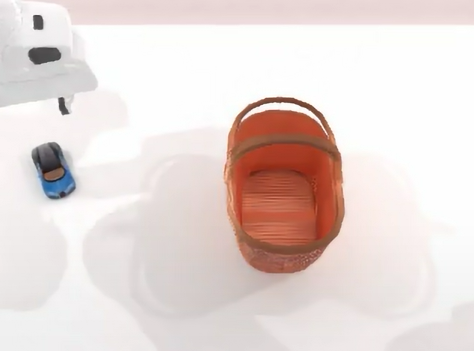}

\caption{Manipulation tasks used in our experiments.
(a) Dynamic pick-and-place in DOM.
(b) Adjust Bottle in DOMINO.
(c) Place Object into Basket in DOMINO.}
\label{fig:task_examples}

\end{figure*}

\paragraph{DOM.}
We construct a pick-and-place task based on DOM~\cite{xie2026dynamicvla}, with the instruction ``pick up the lemon and place it into the red bowl.'' The initial positions of the lemon and bowl are randomized. The dataset contains 100 static demonstrations and 211 dynamic demonstrations, recorded at $25\,\mathrm{Hz}$. In the dynamic setting, the lemon receives an initial velocity uniformly sampled from $[0.05,0.25]\,\mathrm{m/s}$. Actions are represented in Cartesian space. 

\paragraph{DOMINO.}
We use the \emph{Adjust Bottle} and \emph{Place Object Basket} tasks from DOMINO~\cite{fang2026domino} with the dual-arm Franka Panda embodiment. The simulator operates at $30\,\mathrm{Hz}$ and uses Cartesian actions. Each task contains three dynamic levels, with 50 clean and 200 randomized demonstrations per level. This gives 750 demonstrations per task and 1,500 across both tasks. Randomization varies background textures, lighting, table height, and tabletop clutter. Evaluation uses the clean environment at each dynamic level.

\subsection{Real-Robot Tasks and Demonstrations}
\begin{figure*}[!t]
\centering

\includegraphics[width=0.24\textwidth]{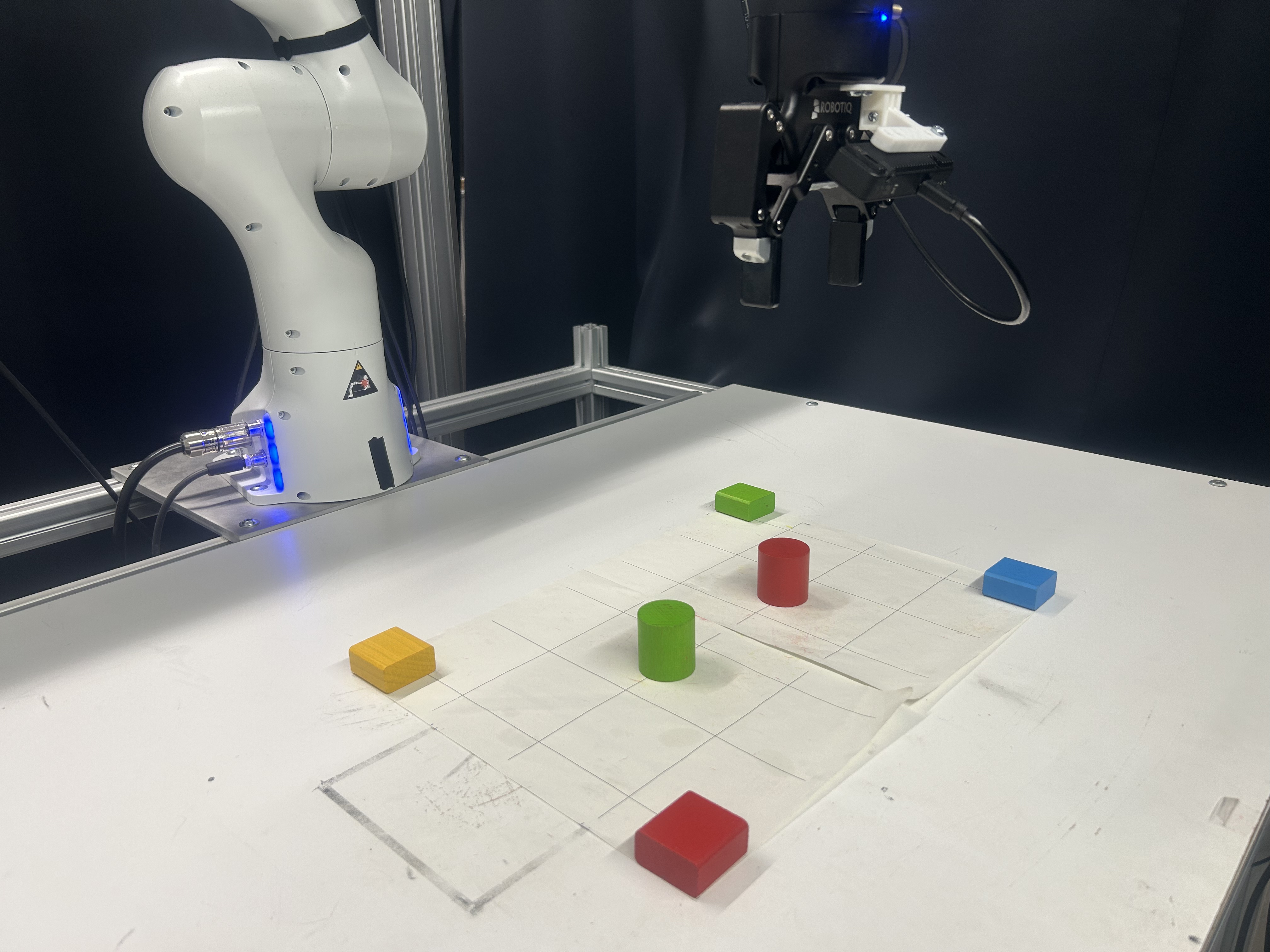}
\hfill
\includegraphics[width=0.24\textwidth]{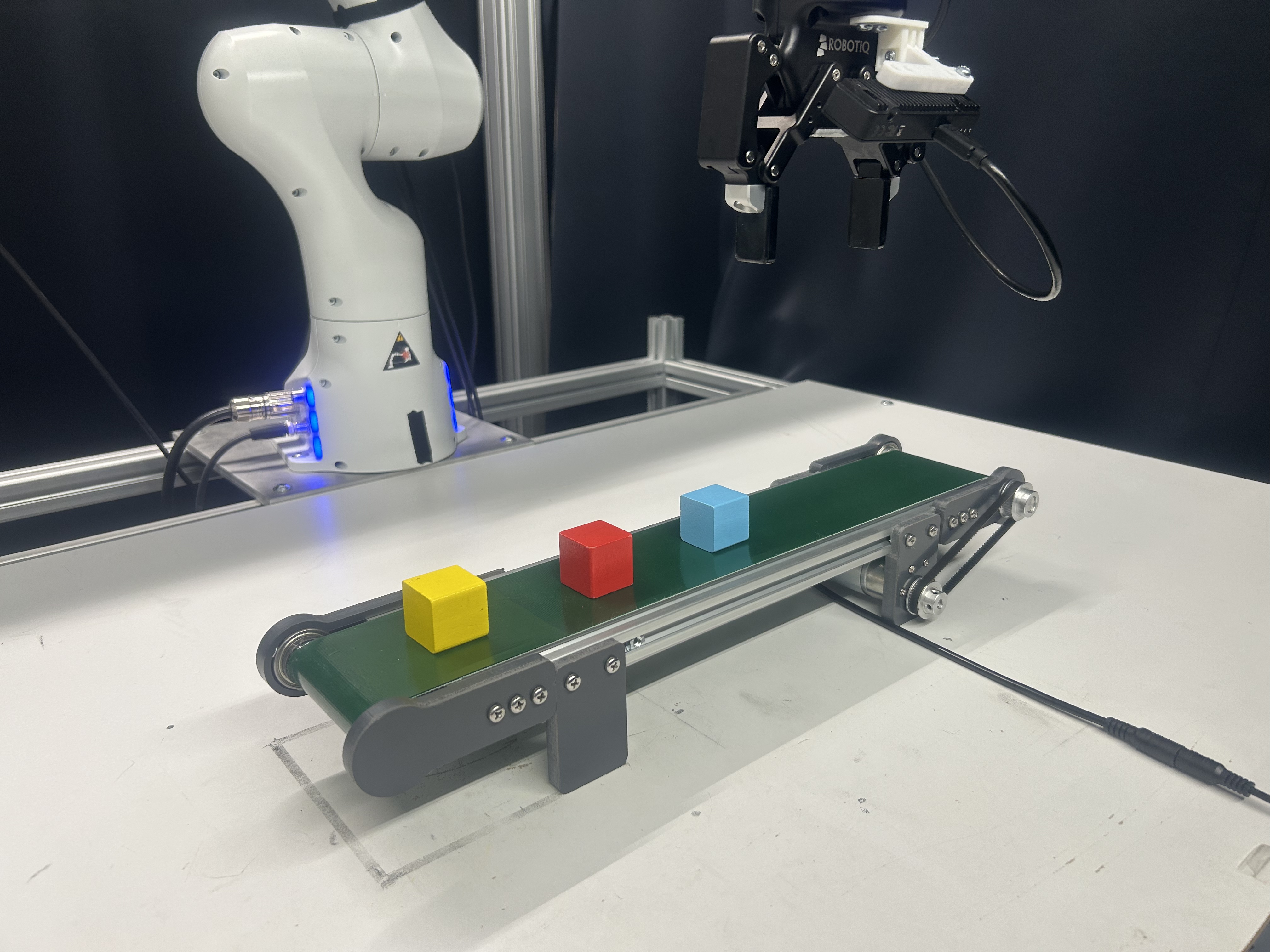}
\hfill
\includegraphics[width=0.24\textwidth]{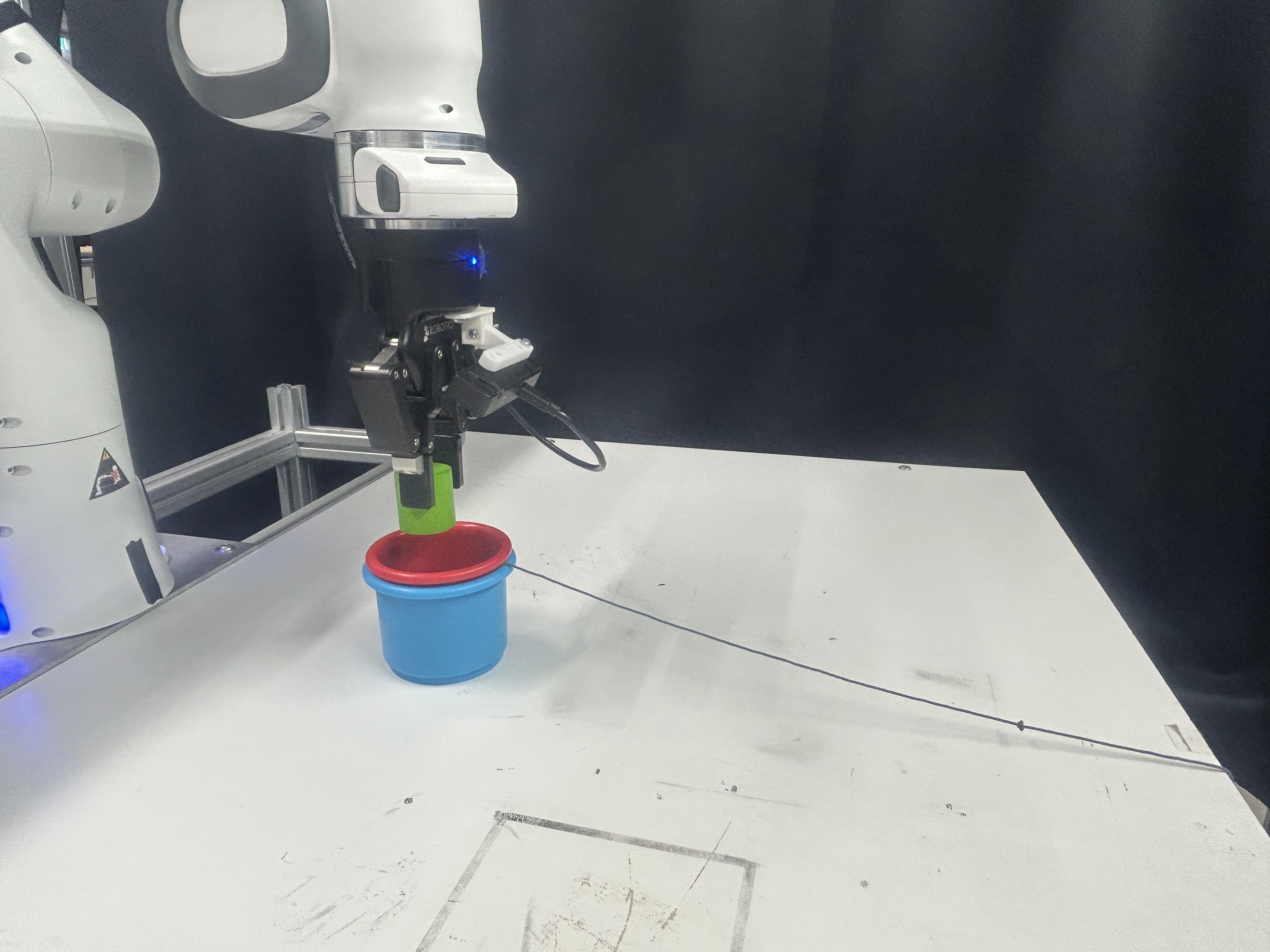}
\hfill
\includegraphics[width=0.24\textwidth]{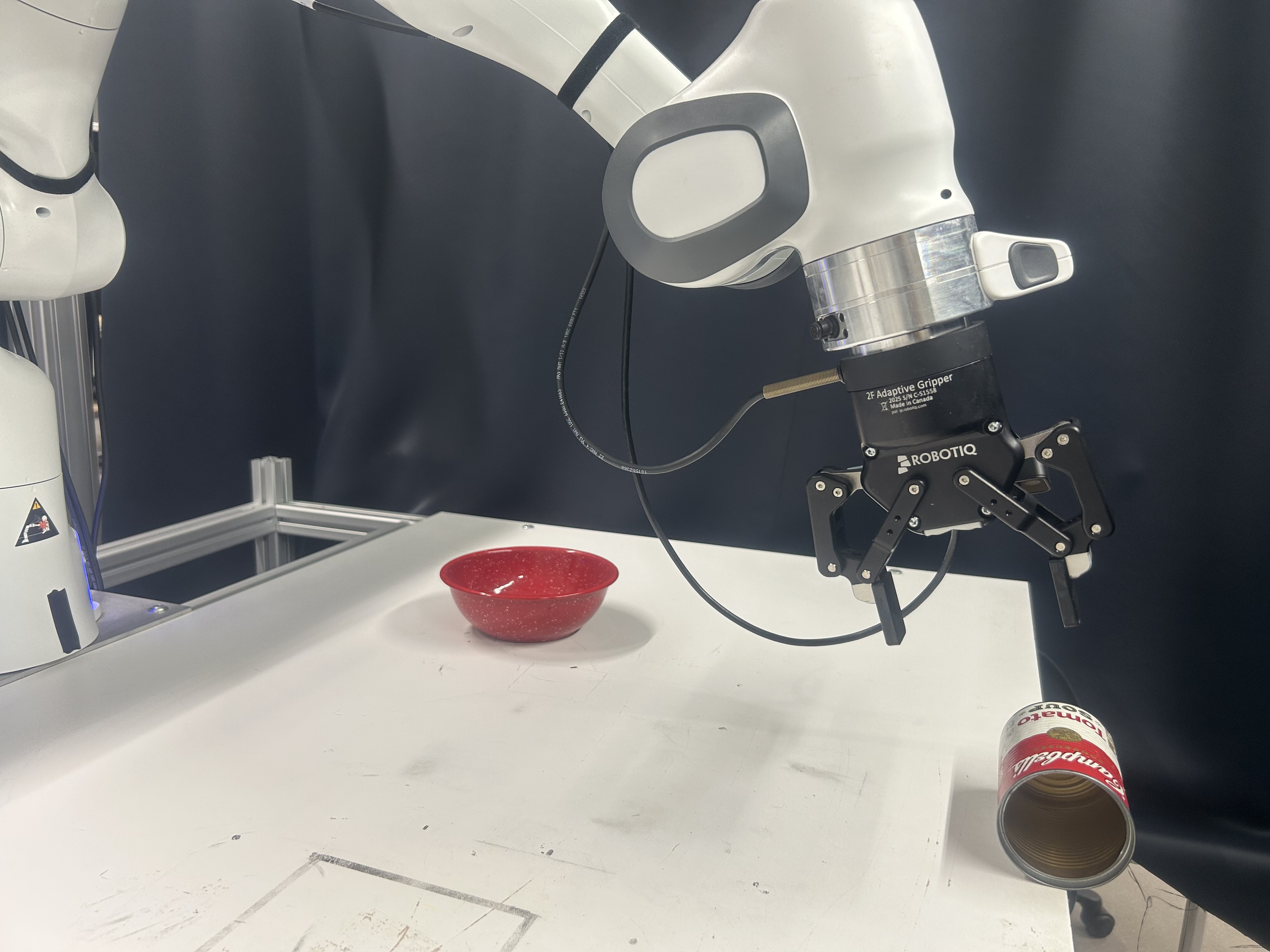}

\caption{Real-robot manipulation tasks.
(a) Static cylinder placement.
(b) Conveyor-belt picking.
(c) Moving-cup placement.
(d) Rolling-can grasping.}
\label{fig:real_robot_tasks}

\end{figure*}
We conduct four real-robot manipulation tasks, including one static task and three dynamic tasks. All demonstrations are collected using a Franka Panda with a Robotiq gripper, a fixed third-person camera, and a wrist-mounted camera. 

\paragraph{Static Cylinder Placement.}
The robot places a specified colored cylinder on a specified colored cube. Four cubes occupy fixed positions, while red and green cylinders vary within a central workspace region. The dataset contains 400 demonstrations, with 50 for each of eight language instructions specifying the cylinder and destination cube.

\paragraph{Conveyor Belt.}
The robot must grasp the red cube from a belt carrying three differently colored cubes and place it into a bowl. The dataset contains 60 demonstrations: 20 with a stationary belt, 20 at a slow speed, and 20 at a medium speed.

\paragraph{Moving Cup.}
The robot starts with a cylinder in its gripper and must place it into a cup. The dataset contains 40 static demonstrations with the cup at different positions and 75 dynamic demonstrations in which an operator pulls the cup using a rope. Pulling speed and direction vary across demonstrations.

\paragraph{Rolling Can.}
An operator rolls a can toward the robot, which must grasp it and place it into a bowl. The dataset contains 60 demonstrations collected at different rolling speeds.

\subsection{Inference and Robot Control}

The policy server runs on a remote A100 GPU node or an NVIDIA GeForce RTX 5090 workstation. The local robot client constructs observations, requests predictions, and buffers returned actions. Predicted actions contain absolute Cartesian positions, quaternion orientations, and gripper commands. Demonstrations are recorded at $20\,\mathrm{Hz}$, while action execution frequencies of $20\,\mathrm{Hz}$ and $30\,\mathrm{Hz}$ are evaluated to investigate their influence on task performance.

The control layer runs at $100\,\mathrm{Hz}$. It reads the latest Cartesian target, limits excessively large position changes, and sends pose commands to the robot. Gripper commands are sent when the requested gripper state changes.

\subsection{Real-Robot Evaluation Protocol}

\paragraph{Method Comparisons.}
For the general method comparisons, each method is evaluated over 20 trials per task. Conveyor-belt trials vary belt speed, cube position, and cube ordering; stationary-belt trials are excluded from this comparison. Rolling-can evaluation contains ten slower and ten faster trials. Moving-cup evaluation varies movement speed and direction. The can and cup are moved manually, so their trajectories cannot be reproduced exactly across methods.

\paragraph{Conveyor-Belt Task Setup and Evaluation.}
We consider three conveyor-belt conditions: stationary,
slow ($2.5\,\mathrm{cm/s}$), and fast ($5\,\mathrm{cm/s}$).
The task involves three cubes, with the red cube as
the target.
In the stationary condition, the target cube can be
placed at three positions along the conveyor belt
(Left, Middle, and Right).
In the dynamic conditions, the belt moves from left
to right, and the cubes initially start on the left side.
Here, Left, Middle, and Right refer to the red cube's
relative position within the three-cube sequence.

For selected experiments, we evaluate nine conditions
formed by three target positions and three belt conditions,
with five trials per condition, giving 45 trials
per configuration.

\paragraph{Targeted DSRL Evaluation.}
The original policy is evaluated over five trials per condition across the nine conveyor-belt conditions. After adaptation, the targeted condition receives 15 trials, while each of the eight remaining conditions receives five trials. We report success counts to make the unequal evaluation sizes explicit. When reporting a mean across conditions, we average the nine condition-specific success rates with equal weight. A pooled success rate instead weights conditions by their trial counts.

We evaluate noise levels of 0.5 and 2.5 for 5k checkpoints, but only 0.5 for 1k checkpoints, as higher noise induces random robot motions.

\subsection{Execution Metrics}
\label{app:execution_metrics}

\paragraph{Time to First Action.}
TTFA is the interval between sending an observation to the policy server and receiving the first executable action:
\begin{equation*}
    \mathrm{TTFA}=t_{\mathrm{first}}-t_{\mathrm{request}}.
\end{equation*}
It includes inference and communication latency, as well as any waiting time imposed by the environment's update cycle within this request--response interval, and measures how soon an executable action from a new chunk becomes available.
For streaming methods, the first action can arrive before generation of the full chunk is complete.

\paragraph{Task Completion Time.}
For successful trials, task completion time is measured from the start of inference to the first gripper release.

\paragraph{Inference Calls.}
Inference calls count the policy requests issued during a trial. It reflects the executed portion of the prediction, which may be shorter than the full prediction horizon.

In the real-robot method comparison, timing and inference-count statistics are computed over successful trials. Success rate uses all evaluation trials.

\section{Supplementary Experimental Results}
\label{app:supplementary_results}

\subsection{Real-Robot Inference and Execution Metrics}
\label{app:inference_metrics}

\begin{table}[ht!]
\centering
\caption{Inference calls and timing metrics for the
real-robot static cylinder placement.}
\begin{tabular}{lcccc}
\toprule
Setting & $s$ & IC & TTFA & TCT \\

\midrule
\multirow{3}{*}{$\pi_{0.5}$ Synchronous}
& 20 & 7.0 & 0.295 s & 8.264 s \\
& 35 & 4.0 & 0.296 s & 7.378 s \\
& 50 & 3.0 & 0.261 s & 6.991 s \\

\midrule
\multirow{3}{*}{$\pi_{0.5}$ with FASTER}
& 5 & 27.5 & \textbf{0.096 s} & 6.728 s \\
& 25 & 6.0 & 0.101 s & 6.608 s \\
& 40 & 4.0 & 0.108 s & 6.709 s \\

\midrule
\multirow{3}{*}{$\pi_{0.5}$ with DynamicVLA}
& 10 & 8.5 & 0.250 s & \textbf{6.019 s} \\
& 20 & 6.5 & 0.260 s & 6.280 s \\
& 30 & 4.5 & 0.295 s & 6.316 s \\

\midrule
\multirow{3}{*}{$\pi_{0.5}$ with RTC}
& 10 & 13.8 & 0.339 s & 6.853 s \\
& 25 & 6.0 & 0.331 s & 6.579 s \\
& 40 & 4.0 & 0.322 s & 6.588 s \\

\midrule
$\pi_{0.5}$ with ABPolicy
& 6 & 20.4 & 0.244 s & 6.754 s \\

\midrule
\multirow{3}{*}{$\pi_{0.5}$ with PUMA}
& 15 & 9.7 & 0.320 s & 9.584 s \\
& 35 & 4.0 & 0.359 s & 7.685 s \\
& 40 & 3.0 & 0.343 s & 7.372 s \\

\bottomrule
\end{tabular}
\label{tab:ttfa_hardware}
\end{table}
Table~\ref{tab:ttfa_hardware} reports inference and execution metrics for the static cylinder placement, with policy inference performed on an NVIDIA A100 GPU and a fixed prediction horizon of $H=50$(Except ABPolicy, $H=32$). We compare synchronous $\pi_{0.5}$ with FASTER, DynamicVLA, RTC, ABPolicy, and PUMA. Execution horizons are varied where applicable to evaluate their effects on inference calls, time to first action, and task completion time.

\begin{table}[ht!]
\centering
\caption{Inference calls and timing metrics for
asynchronous $\pi_{0.5}$ on the dynamic conveyor-belt task.}
\label{tab:5090_metrics}
\begin{tabular}{lcccc}
\toprule
Setting & $s$ & IC & TTFA & TCT \\
\midrule

\multirow{3}{*}{$\pi_{0.5}$ Asynchronous}
& 5  & 61.0 & 0.145 s & 15.088 s \\
& 10 & 22.5 & 0.144 s & 10.952 s \\
& 15 & 12.1 & 0.143 s & 8.624 s \\

\midrule
$\pi_{0.5}$ Asynchronous + AAC
& Adaptive & 12.1 & 0.285 s & 7.436 s \\

\bottomrule
\end{tabular}
\end{table}
Table~\ref{tab:5090_metrics} reports the same metrics for the dynamic conveyor-belt task, with inference performed on an NVIDIA RTX 5090 GPU. We evaluate asynchronous $\pi_{0.5}$ using the 5k checkpoint and a fixed prediction horizon of $H=20$, comparing fixed execution horizons ($s=5,10,15$) with Adaptive Action Chunking (AAC). Timing and inference-count metrics are averaged over successful trials.


\end{document}